\documentclass[twocolumn]{article}
\usepackage[utf8]{inputenc}
\usepackage{adjustbox}
\usepackage[english]{babel}

\usepackage{comment}             
\usepackage{fancyhdr, lastpage}  
\usepackage{float}               
\usepackage{titlesec}            
\usepackage{booktabs}            
\usepackage{colortbl}            
\usepackage{subcaption}          
\usepackage[T1]{fontenc}         
\usepackage{tabularx}
\usepackage{chngcntr}            
\usepackage[hidelinks]{hyperref} 

\usepackage{ragged2e} 

\usepackage[style=nejm,backend=biber]{biblatex}

\renewrobustcmd*{\bibinitperiod}{\addspace}
\usepackage{csquotes}

\usepackage{glossaries} 

\usepackage{graphicx}   
\usepackage{subcaption} 

\usepackage{verbatim}   
\usepackage{mathptmx}   
\usepackage{cleveref}   

\usepackage[left=2.74cm, right=2.74cm, top=3cm]{geometry}
\usepackage[font=small]{caption}                         
\usepackage[affil-it]{authblk}

\title{Automated voice- and text-based classification of neuropsychiatric conditions in a multidiagnostic setting}
\author[1,2,3]{Lasse Hansen\thanks{lasse.hansen@clin.au.dk}}
\author[4]{Roberta Rocca}
\author[4,5]{Arndis Simonsen}
\author[4,6]{Alberto Parola}
\author[2,4,5]{Vibeke Bliksted}
\author[1,2]{Nicolai Ladegaard}
\author[7,8,9]{Dan Bang}
\author[4,6]{Kristian Tylén}
\author[4,6]{Ethan Weed}
\author[1,2]{Søren Dinesen Østergaard}
\author[4,6,10]{Riccardo Fusaroli}

\affil[1]{\footnotesize Department of Affective Disorders, Aarhus University Hospital - Psychiatry, Aarhus, Denmark}
\affil[2]{\footnotesize Department of Clinical Medicine, Aarhus University, Aarhus, Denmark}
\affil[3]{\footnotesize Center for Humanities Computing, Aarhus University, Aarhus, Denmark}
\affil[4]{\footnotesize Interacting Minds Centre, Aarhus University, Aarhus, Denmark}
\affil[5]{\footnotesize Psychosis Research Unit, Aarhus University Hospital - Psychiatry, Aarhus, Denmark}
\affil[6]{\footnotesize Department of Linguistics, Cognitive Science and Semiotics, Aarhus University, Aarhus, Denmark}
\affil[7]{\footnotesize Wellcome Centre for Human Neuroimaging, University College London, WC1N 3BG London, United Kingdom}
\affil[8]{\footnotesize Department of Experimental Psychology, University of Oxford, OX2 6GG Oxford, United Kingdom}
\affil[9]{\footnotesize Center of Functionally Integrative Neuroscience, Aarhus University, Aarhus, Denmark}
\affil[10]{\footnotesize Linguistic Data Consortium, University of Pennsylvania, Philadelphia, Pennsylvania, USA}

\makeglossaries
\newacronym{asd}{ASD}{Autism Spectrum Disorder}
\newacronym{mfccs}{MFCCs}{Mel-Frequency Cepstral Coefficients}
\newacronym{nlp}{NLP}{Natural Language Processing}
\newacronym{cv}{CV}{Cross-validation}
\newacronym{mdd}{MDD}{Major Depressive Disorder}
\newacronym{tfidf}{TF-IDF}{Term Frequency–Inverse Document Frequency}
\newacronym{cnn}{CNN}{Convolutional Neural Network}
\newacronym{schz}{SCHZ}{Schizophrenia}

\begin{document}
\raggedbottom

\maketitle
   {\noindent

\textbf{Abstract:} Speech patterns have been identified as potential diagnostic markers for neuropsychiatric conditions. However, most studies only compare a single clinical group to healthy controls, whereas clinical practice often requires differentiating between multiple potential diagnoses (multiclass setting). To address this, we assembled a dataset of repeated recordings from 420 participants (67 with major depressive disorder (MDD), 106 with schizophrenia and 46 with autism spectrum disorder (ASD), as well as matched controls), and tested the performance of conventional machine learning models and advanced Transformer models on both binary and multiclass classification, based on voice and text features. 

While binary models performed comparably to previous research (F1 scores between 0.54-0.75 for ASD; 0.67-0.92 for MDD; and 0.71-0.83 for schizophrenia); when differentiating between multiple diagnostic groups performance decreased markedly (F1 scores between 0.35-0.44 for ASD, 0.57-0.75 for MDD, 0.15-0.66 for schizophrenia, and 0.38-0.52 macro F1). Combining voice and text-based models increased performance, suggesting that they capture complementary diagnostic information.

Our results indicate that models trained on binary classification might rely on markers of \textit{generic} differences between clinical and non-clinical populations, or markers of clinical features that overlap across conditions, rather than identifying markers \textit{specific} to individual conditions. We provide recommendations for future research in the field, suggesting increased focus on developing larger transdiagnostic datasets that include more fine-grained clinical features, and that can support the development of models that better capture the complexity of neuropsychiatric conditions and naturalistic diagnostic assessment. 
}
\vspace{3mm}
\hrule
\vspace{-2mm}

\section{Introduction} 
\label{sec:introduction}

Machine learning is increasingly used to identify voice and text-based markers that complement current methods for screening and monitoring neuropsychiatric conditions \parencite{macfarlane_combining_2022, he_deep_2022}. However, while clinical assessment often requires the identification of one diagnosis amongst many with partially overlapping clinical features, virtually all machine learning studies of voice and text-based markers focus on simpler binary outcomes, i.e., patients with one specific diagnosis \textit{versus} controls without any neuropsychiatric conditions \parencite{parola_voice_2020, low_automated_2020, koops_speech_2023, fusaroli_is_2017, hansen_generalizable_2022}. This raises concerns that the markers identified by studies in one-diagnosis settings may not have diagnostic specificity -- instead capturing general differences between clinical and non-clinical populations. 
 
In this paper, we briefly review the rationale behind using voice and text-based markers of neuropsychiatric disorders, and the techniques available to do so (including the most recent deep learning Transformer model). We then investigate their potential to discriminate between \textit{multiple} diagnoses at the same time -- a more challenging, but more clinically relevant task than deciding if a person is best classified with one particular psychiatric diagnosis or not -- using a novel speech dataset including patients with autism spectrum disorder (ASD), major depressive disorder (MDD) and schizophrenia.

\subsection{Previous work on vocal and linguistic markers of neuropsychiatric conditions}

Early descriptions of individuals with neuropsychiatric conditions tend to include remarks about atypical communicative patterns \parencite{kraepelin_manic-depressive_1921, hamilton_rating_1960}. Individuals with ASD are often described as producing stereotyped or robotic-like speech \parencite{fusaroli_is_2017, fusaroli_toward_2022, rybner_vocal_2022}, individuals with schizophrenia tend to produce more monotone, fragmented and less coherent speech \parencite{parola_speech_2022, parola_voice_2022}, and depression has been associated with producing sentences with longer pauses and more negative affect \parencite{cummins_review_2015, koops_speech_2023, nguyen_affective_2014}. Accordingly, variations in what people say and how they say it are directly or indirectly used as part of the diagnostic criteria for mood disorders, schizophrenia, and ASD \parencite{world_health_organization_icd-10_1993, association_diagnostic_2013}. 

Traditionally, voice-based classification of neuropsychiatric conditions has relied on intuitive handcrafted features that were easy to measure, such as pitch, number and duration of pauses, and response latency \parencite{jensen_metavoice_2022}. Today, advances in natural language processing (NLP) and speech processing methodologies allow for largely automated extraction of a broader range of features, such as formants, energy, and spectral and cepstral descriptors (e.g., mel-frequency cepstral coefficients (MFCCs)) from audio, relying on standard software libraries such as OpenSMILE and Covarep \parencite{eyben_opensmile::_2015, degottex_covarepcollaborative_2014}. 
Texts and transcribed speech can also be quantified in terms of features such as relative frequencies of words, sentiment expressed, and semantic embeddings. These features are motivated by general research on linguistic production and clinical observations of specific patient groups (e.g. longer pauses in individuals with alogia - a symptom of schizophrenia - which can be measured from voice recordings, and more negative emotional words in individuals with depression, which can be measured by applying text-based tools to transcripts). Classifiers relying on these human-defined features have often reported promising performance in discriminating diagnosed individuals (e.g., with schizophrenia) from controls without any neuropsychiatric condition, with accuracies between 60\% and 90\% (see \textcite{fusaroli_is_2017, koops_speech_2023, parola_voice_2020, rybner_vocal_2022} for an overview of existing studies). 
Furthermore, several studies have sought to combine the two modalities (speech, text) and found them to contain complementary information for the classification of neuropsychiatric conditions \parencite{voppel_semantic_2022, macfarlane_combining_2022}. This complementarity may be due to the ability of text models to capture variations in word usage that are not captured by acoustic models, and the ability of acoustic models to identify variations in prosody (e.g. tone, pitch, or rhythm) that are not expressed in text.

Recent approaches to the encoding of text and audio, which use deep neural networks and self-attention \parencite{vaswani_attention_2017}, have eschewed traditional feature engineering \parencite{lecun_deep_2015, cummins_speech_2018}. Briefly, self-attention allows the model to selectively focus on different parts of the input sequence, enabling it to effectively incorporate contextual information from the entire input into its predictions \parencite{vaswani_attention_2017}. Models like Transformers, which leverage self-attention to produce encodings of words and acoustic patterns modulated by the context surrounding them \parencite{vaswani_attention_2017}, outperform other approaches across fields ranging from automated speech recognition to natural language understanding \parencite{baevski_wav2vec_2020, devlin_bert_2019}. These models are generally pre-trained on generic language tasks using large amounts of data, and can be deployed with minimal additional fine-tuning on novel data and novel predictive tasks. This approach, known as \textit{transfer learning}, has been highly successful in many natural language processing tasks \parencite{ruder_transfer_2019}, but it has not been widely tested in the context of neuropsychiatric assessments based on vocal and text markers \parencite{hansen_generalizable_2022}. 

\subsection{Multidiagnostic challenges}\label{multidiag:sec:clinical_utility} 

Previous attempts at identifying voice and text-based markers of neuropsychiatric conditions have primarily focused on clear binary contrasts: given a set of clearly diagnosed individuals, could voice markers discriminate them from controls with no symptoms or diagnoses \parencite{low_automated_2020, koops_speech_2023}? Yet, clinical practice is much more likely to involve discrimination between multiple diagnoses with partially overlapping clinical features \parencite{world_health_organization_icd-10_1993, association_diagnostic_2013}. This more clinically relevant task is also likely to be a much more challenging classification since some vocal markers are shared across conditions (e.g. decreased pitch variability in both MDD and schizophrenia; or longer pauses in both ASD and schizophrenia) \parencite{jensen_metavoice_2022, parola_speech_2022, koops_speech_2023}. 

\subsection{Aims of the study}\label{multidiag:sec:research_statement}
Given the clinical limitations of binary classification, this study aimed to evaluate the performance of voice- and text-based machine learning models of neuropsychiatric conditions in multiclass classification settings. Compared to traditional binary classification tasks, which contrast diagnosed individuals with matched controls, multiclass classification better mirrors the complexity of clinical assessment and provides a better understanding of the potential clinical utility of machine learning models. Therefore, we established a multidiagnosis dataset consisting of individuals with MDD, ASD and depression to answer the following research questions:

\begin{itemize}
    \item How well can we distinguish individuals with MDD, ASD, or schizophrenia from matched controls based on speech patterns in a binary classification setting?
    \item How good are traditional machine learning models at identifying \textit{diagnosis-specific} speech-based markers? In other words, how does multi-class performance compare to binary performance?
    \item How do voice- \textit{vs} text-based models compare? Does combining the two modalities yield better predictions?
    \item How do pre-trained Transformer models perform compared to more traditional feature-based models?
\end{itemize}

\section{Methods} 
\label{sec:methods}
In this study, we trained binary and multi-class models on voice and transcripts from Danish adult individuals with a diagnosis of schizophrenia, MDD, or ASD as well as healthy matched controls performing the Animated Triangles (Frith-Happé) task \parencite{abell_triangles_2000}. We first assessed whether we can achieve binary classification performance comparable to existing studies. We then evaluated how well the same architectures performed in a multi-class setup. In order to extensively assess these issues, we used both voice recordings and text-based properties of transcribed speech, and contrasted more traditional machine learning approaches with pre-trained state-of-the-art Transformer models. The following sections provide details on the data and modelling strategies. 

\subsection{Pipeline}
\begin{figure*}[]
    \centering
    \includegraphics[width=0.85\textwidth]{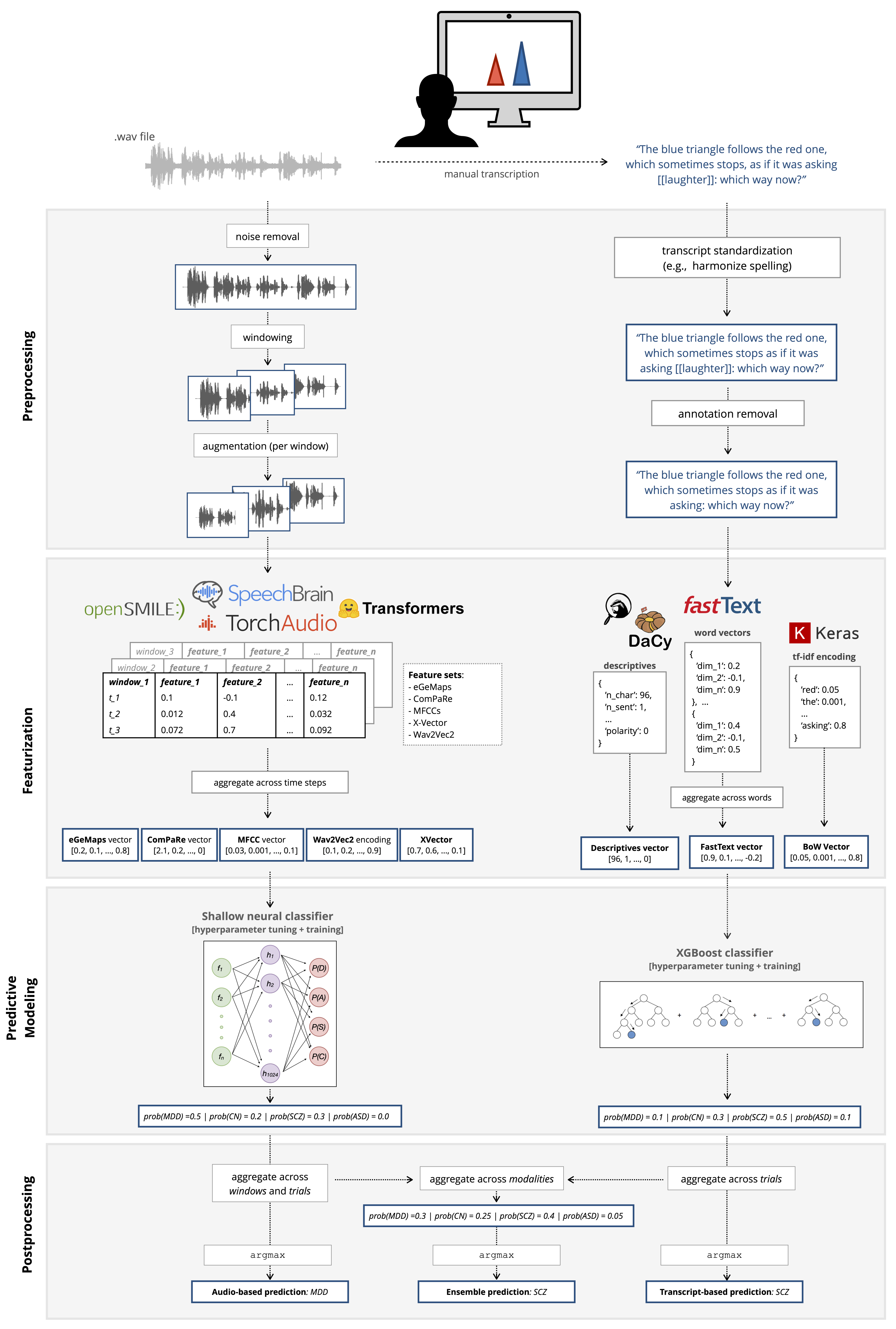}
    \caption{Processing and training pipelines for feature-based text, voice, and ensemble models (see \autoref{sec:methods} and \autoref{sec:appendix} for details on each processing step). For transformer-based text models, we use the same pre- and postprocessing step, while feature extraction and predictive modeling are performed end-to-end within the transformer architecture.}
    \label{fig:pipeline}
\end{figure*}

For training, we split the data into a training set, a validation set, and a balanced held-out test set. We ensure that the models are trained (fitted) and optimized on one subset of the data (the training + validation dataset), while their performance is assessed on a different subset of the data (the held-out test dataset). \autoref{fig:pipeline} provides an overview of the full pipeline, which is discussed in detail below.

\begin{table*}
\centering
\caption{Overview of demographic and clinical information available for each study.}
\label{tab:table1}
\begin{tabular}[t]{lllllll}
\toprule
\multicolumn{1}{c}{ } & \multicolumn{2}{c}{Autism Spectrum Disorder} & \multicolumn{2}{c}{Major Depressive Disorder} & \multicolumn{2}{c}{Schizophrenia} \\
\cmidrule(l{3pt}r{3pt}){2-3} \cmidrule(l{3pt}r{3pt}){4-5} \cmidrule(l{3pt}r{3pt}){6-7}
  & ASD & Controls & MDD & Controls & SCZ & Controls\\
\midrule
 & (N=46) & (N=41) & (N=67) & (N=42) & (N=106) & (N=118)\\
\addlinespace[0.3em]
\multicolumn{7}{l}{\textbf{Age}}\\
\hspace{1em}Mean (SD) & 28.0 (7.23) & 26.1 (6.42) & 34.5 (11.1) & 33.5 (12.0) & 27.1 (9.79) & 27.1 (9.78)\\
\addlinespace[0.3em]
\multicolumn{7}{l}{\textbf{Sex}}\\
\hspace{1em}Female & 19 (41.3\%) & 17 (41.5\%) & 50 (74.6\%) & 30 (71.4\%) & 44 (41.5\%) & 50 (42.4\%)\\
\hspace{1em}Male & 26 (56.5\%) & 24 (58.5\%) & 15 (22.4\%) & 11 (26.2\%) & 60 (56.6\%) & 66 (55.9\%)\\
\addlinespace[0.3em]
\multicolumn{7}{l}{\textbf{Education (years)}}\\
\hspace{1em}Mean (SD) & 12.2 (3.40) & 13.8 (2.82) & 12.7 (2.12) & 12.9 (1.97) & 15.2 (2.72) & 14.9 (2.52)\\
\addlinespace[0.3em]
\multicolumn{7}{l}{\textbf{HDRS-17}}\\
\hspace{1em}Mean (SD) & - & - & 22.5 (3.71) & 1.62 (1.36) & - & -\\
\addlinespace[0.3em]
\multicolumn{7}{l}{\textbf{ADOS}}\\
\hspace{1em}Mean (SD) & 32.5 (7.03) & 13.8 (9.05) & - & - & - & - \\
\multicolumn{7}{l}{\textbf{Uses antipsychotics}}\\
\hspace{1em}Yes & - & - & - & - & 0 (0\%) & 28 (26.4\%)\\
\bottomrule
\multicolumn{7}{l}{\rule{0pt}{1em}Top-level grouping specifies which study the participants come from.}\\
\multicolumn{7}{l}{\rule{0pt}{1em}Note that counts for sex might not add up to the total number of participants due to incomplete data.}\\
\multicolumn{7}{l}{\rule{0pt}{1em}HDRS-17: The total score on the 17-item Hamilton Depression Rating Scale.}\\
\multicolumn{7}{l}{\rule{0pt}{1em}ADOS: The total score on the Autism Diagnostic Observation Schedule.}\\

\end{tabular}
\end{table*}

\subsection{Data}
The dataset consists of speech recordings and corresponding word-level transcriptions from 420 individuals with MDD (N=67), ASD (N=46), and schizophrenia (N=106) along with controls matched on sex, age and socio-economic status (N=201). The data were collected in previous studies for other purposes \parencite{ladegaard_comparison_2014, ladegaard_higher-order_2014, beck_cross-cultural_2020, bliksted_hyper-and_2019, bliksted_social_2014}. For demographic, clinical, and cognitive information about the participants, see \autoref{tab:table1}. See Appendix \cref{sec:appendix:original_studies} for a description of the original studies. 

All participants performed the Animated Triangles task \parencite{abell_triangles_2000}: they watched 8-10 short videos of geometric figures moving on the screen and after each video produced a description of it (for more details on the task, see \autoref{sec:appendix:triangles}). The task was chosen as it generates relatively free speech production - shown to produce more distinctive voice patterns in patient populations than, for instance, reading aloud a text \parencite{cummins_review_2015, parola_voice_2020, sechidis_machine_2021} - but still pertaining to a common domain (e.g., all will mention triangles).

Initial data collection was approved by the IRB at the Central Denmark Region (reference numbers 2007-58-0010, 2009-0035, 2010-0161, 264/2014), which also deemed the current analysis exempt from the need of further ethical approval.

\autoref{app_tab:audio_description} and \autoref{app_tab:transcript_description} in the Appendix present the number of participants and size of the audio and transcript data, respectively. Some participants did not complete all animations and some recordings were excluded due to poor audio quality.

\subsubsection{Data collection and preprocessing}
Recording equipment and procedures were consistent across studies: quiet but not soundproof rooms, and tabletop microphones. To prepare the recordings for analysis, we manually identified and fixed issues with sound quality, tagged segments with participant speech, and transcribed them after removing background noise. Recordings with fewer than 3 seconds of participant speech were removed, and the remaining ones were split into 5-second segments with a 1-second sliding window. Transcripts were manually checked for errors and inconsistencies. See \autoref{fig:pipeline}(B) for a visualization, and \autoref{sec:appendix:preprocessing} for more details on preprocessing.

\subsubsection{Data Splits}
Validation and test sets - on which we test the performance of the machine learning models - were constructed to equally represent each diagnosis and sex. Both validation and test set included 36 participants (for each diagnostic group, 6 randomly chosen participants with diagnosis and 6 matched controls). The training data included the remaining participants. No participant was included in multiple splits. 

\subsection{Models}
For voice and text separately, we created two types of predictive models: models relying on more "traditional" machine learning pipelines (henceforth: baseline models) and Transformer models with added classification heads (henceforth: Transformer models). Each model was trained on three binary classification tasks (for each diagnosis, predicting whether or not the participant is diagnosed with the condition) and on multi-class classification, that is, on predicting the correct diagnostic group (MDD, ASD, schizophrenia, or control) in a setup where data from all other diagnostic groups and their matched controls are included. 

\subsubsection{Baseline Models}

We created four baseline audio models using different sets of features as inputs, respectively relying on the eGeMAPS and ComParE feature sets (\textit{openSMILE} Python library v2.4.1 \parencite{eyben_opensmile::_2015}), 40 MFCCs (\textit{torchaudio} v0.11.0), and representations from the final layer of a non-contextual deep learning model pre-trained on speaker recognition (X-Vector, \parencite{snyder_x-vectors_2018}  \parencite{speechbrain_speechbrainspkrec-xvect-voxceleb_2022}).
Baseline audio models consisted of shallow classifiers with one intermediate layer of size 1024, with input dimensionality defined by the number of input features (88 for eGeMAPS, 6373 for ComParE, 40 for MFCCs, and 512 for X-Vector).

For text, we trained 28 baseline models: one for each of the basic statistical descriptors defined in \textit{TextDescriptives} v.1.0.7 Python library \parencite{hansen_textdescriptives_2023}; a model including all of these descriptors as input features simultaneously; three TF-IDF models (term frequency - inverse document frequency, \parencite{jones_statistical_1972}) of varying dimensionality (100, 1000 and 10000) implemented in Keras (v.2.7.0); and one model using pre-trained \textit{fasttext} word embeddings \parencite{bojanowski_enriching_2016}. See \autoref{tab:text_features} for an overview of all features and models. Note that, as the focus of the present paper is a comprehensive evaluation of the predictive performance of automatically extractable feature sets in a multiclass vs. binary setting, we do not delve into an in-depth qualitative discussion of predictive performance of individual features.

Text baseline models were trained using XGBoost, which implements regularised gradient boosted decision trees that are robust to heterogeneous feature sets \parencite{chen_xgboost_2015}. XGBoost combines bagging and boosting to achieve optimal bias-variance trade-off, and to reduce overfitting. XGBoost hyperparameters for baseline models were found via randomized grid search over the parameter grid defined in \autoref{tab:xgboost_grid}, and optimized on the validation set. An overview on the baseline models and the features used in each is presented in \autoref{tab:text_features} and \autoref{multidiag:tab:n_features}. 

\subsubsection{Transformer Models}

We relied on 3 pre-trained audio Transformer models - \textit{wav2vec2-base-da}, \parencite{alvenir_alvenirwav2vec2-base-da_2022}, multilingual \textit{XLS-R} \parencite{babu_xls-r_2021}, a Danish version of \textit{XLS-R}, \parencite{hansen_chcaaxls-r-300m-danish_2022} -  and 4 text-based models - \textit{ScandiBERT} \parencite{snaebjarnarson_vesteinnscandibert_2022}, \textit{XLM-RoBERTa-base} \parencite{conneau_unsupervised_2020}, \textit{Multilingual MiniLM Sentence Transformer} \parencite{reimers_making_2020}, and \textit{Danish Electra} (Ælectra) \parencite{hojmark-bertelsen_aelaectra_2021}.

We optimized the hyperparameters for our text-based Transformers and fine-tuned them both by freezing the encoder and only training the classification head and by fine-tuning the whole model (see \autoref{sec:appendix:models}). We then selected the best-performing model from each of the 4 pre-trained model classes by evaluating them on the validation set. Models were fed transcripts from individual trials as input, yielding one prediction per trial. Class probabilities for multiple transcripts from the same individual were aggregated to produce one prediction per individual during evaluation.

For audio models -- which are significantly more computationally demanding -- we froze the weights of the entire model and trained the classification head only. We tuned the learning rate, and selected the best-performing model based on the performance on the evaluation set. As with baseline models, we aggregated the extracted features over each 5-second window of recording using the mean before feeding them to the classification layer. To get a single prediction per recording, we aggregated the model output from all respective 5-second windows using the mean. Full details on models and training procedures (including data augmentation) are provided in the appendix, but see also \autoref{fig:pipeline}.

\begin{figure*}[!ht]
    \centering
    \includegraphics[width=\textwidth]{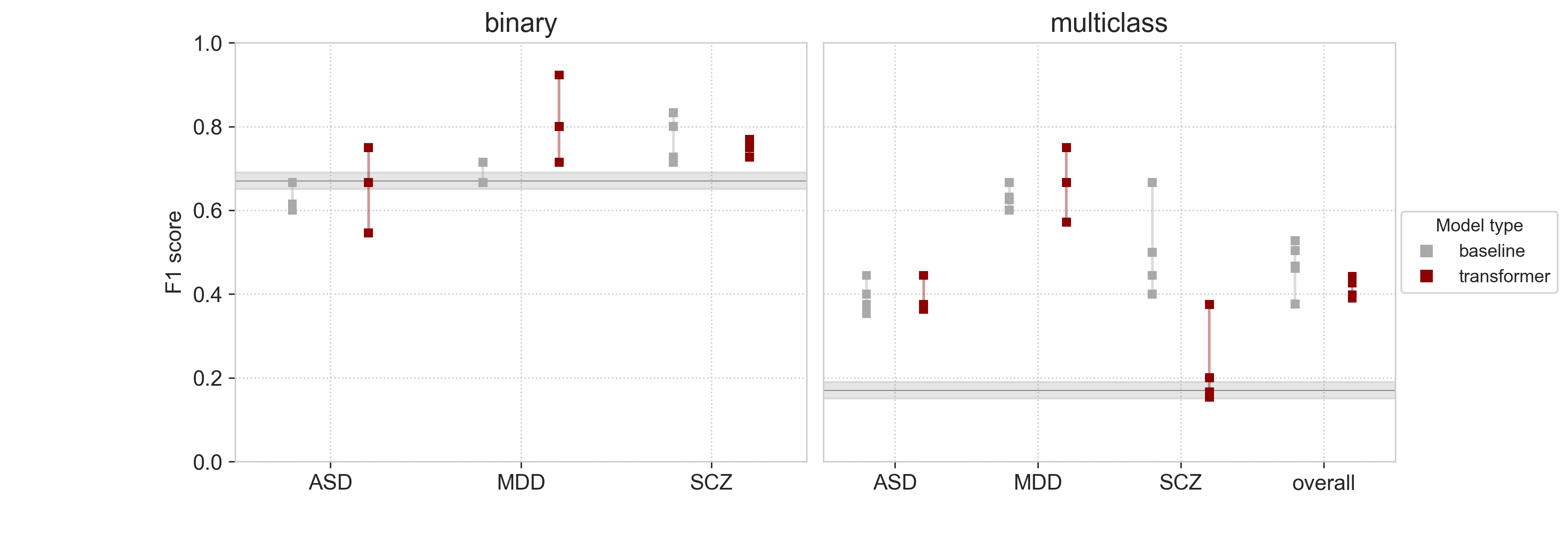}
    \caption{Binary and multiclass performance per group and classification setup. The x-axis indicates the three different diagnostic groups, as well as overall performance in the multiclass setting. The y-axis indicates performance measured in terms of F1-scores for group \textit{vs} matched controls (left panel) or group \textit{vs} all other participants (including other diagnostic groups, right panel). The best 5 baseline models per group are in grey, the best 5 Transformer models in red. The grey horizontal bar indicates the performance of a highly conservative dummy baseline which always predicts "control".}
    \label{fig:multiclass_vs_binary}
\end{figure*}

\subsubsection{Multi-modal Combination}

To estimate whether text and voice models provided overlapping or complementary information, we created ensemble models that combined predictions from the best performing models, in terms of macro F1 on the test set, from each modality. The F1 score is a useful metric for comparison, as it incorporates both the sensitivity and the positive predictive value of a classification model. Macro F1 is obtained by taking the unweighted average of the per-class F1 score. We created two ensemble models, one combining the two best-performing baselines for audio and text, and one combining the best-performing Transformer models. For each trial, the ensemble models selected the output of the most confident model \autoref{fig:pipeline}. Cross-trial aggregation was performed analogously to text- and voice-based models.

\section{Results} 
\label{sec:results}
    \subsection{Binary performance}
When trained on identifying individuals with a diagnosis versus matched controls in a unimodal binary classification setting, the 5 best baseline models and transformer models achieved F1 classification performance of 0.54-0.75 for ASD, 0.67-0.92 for MDD, and 0.71-0.83 for schizophrenia (see \autoref{fig:multiclass_vs_binary}). These ranges are comparable to those reported in previous literature \parencite{fusaroli_is_2017, parola_voice_2020, koops_speech_2023}. The best Transformer models performed better than baseline models for MDD and ASD, and were on par with baselines models for schizophrenia (see \autoref{fig:multiclass_vs_binary}). See \autoref{tab:schz_perf}, \autoref{tab:asd_perf} and \autoref{tab:depr_perf} for further details. 

\subsection{Multi-diagnostic performance}

When trained on predicting diagnoses in a multi-class setting, overall F1 scores (F1-macro) were markedly lower than binary F1 scores, ranging between 0.38 and 0.57 for the top 5 baselines models, and between 0.41 and 0.46 for the top 5 Transformers (see \autoref{fig:multiclass_vs_binary}). Class-specific F1 scores were comparable to binary scores only for MDD, and were lower for ASD and schizophrenia. Transformers' performance was not generally better than baseline models'. Class-specific F1 scores were slightly higher for transformers compared to baselines for MDD, they were lower than baseline models for ASD, and markedly worse for schizophrenia. See \autoref{tab:multiclass_perf} for further details.

\subsection{Voice vs Transcripts}

\begin{figure*}[!ht]
    \centering
    \includegraphics[width=0.85\textwidth, scale=.9]{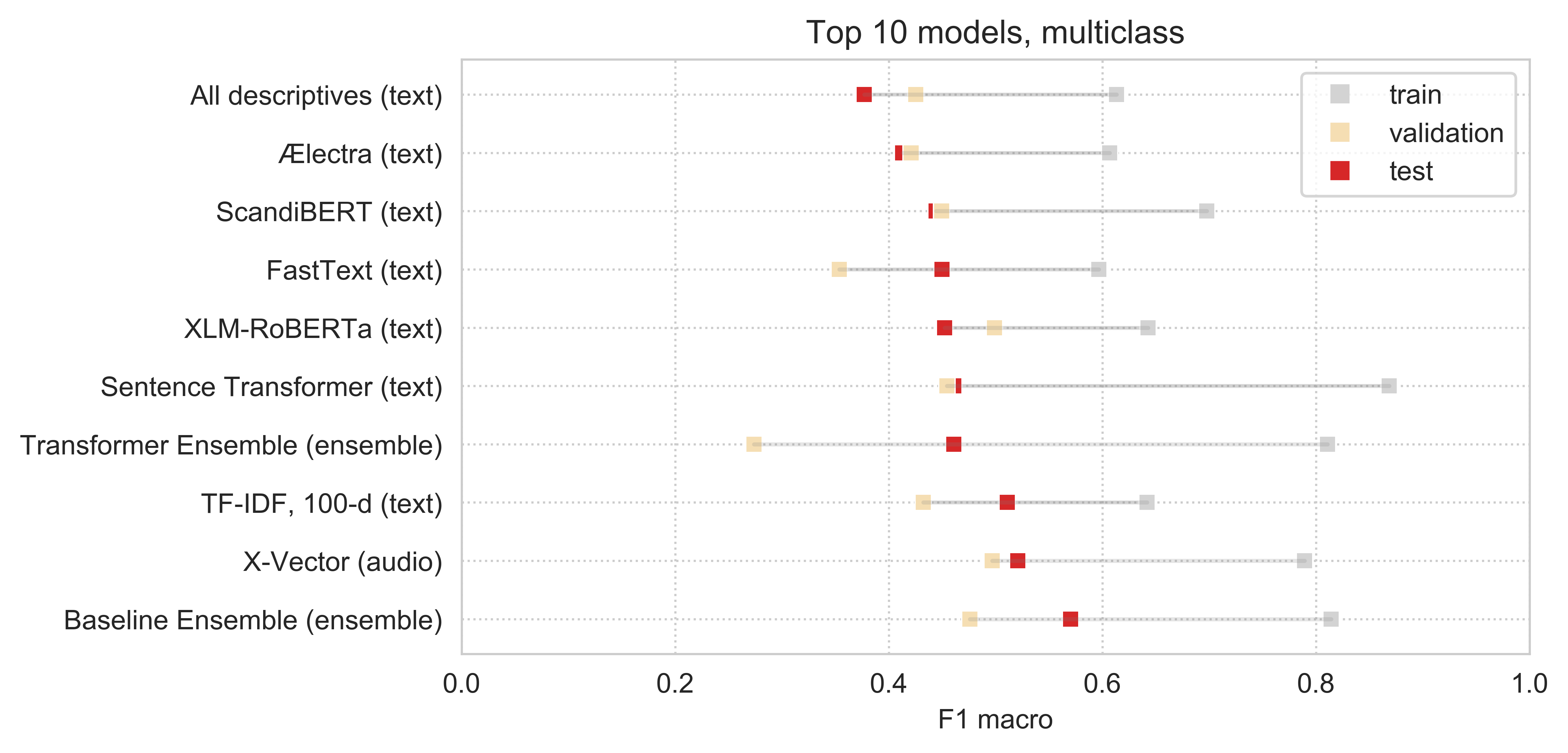}
    \caption{Detailed performance overview for the 10 best multiclass models. Y-axis indicates model details (with models ordered by ascending performance). X-axis indicates performance, quantified as F1 macro, on the the training, validation and test set. Model names on the y-axis are tagged by modality. Large gaps between training and test performance indicates overfitting on the training data.}
    \label{fig:multiclass_top_10}
\end{figure*}

As seen in \autoref{fig:multiclass_top_10}, the best voice-based model (\textit{X-Vector}) and text-based model (\textit{100-dimensional TF-IDF}) performed on par, but they were outperformed by an ensemble model pooling the outputs of the two (\textit{Baseline Ensemble}). Transformers also feature among the top 10 models, but only text ones, and their performance is not better than the best baseline models. Transformer-based ensemble pooling does not improve performance over the best text Transformer.

\subsection{Error Analysis}
\begin{figure*}[!ht]
    \centering
    \includegraphics[width=\textwidth, trim={0 15cm 13cm 0}]{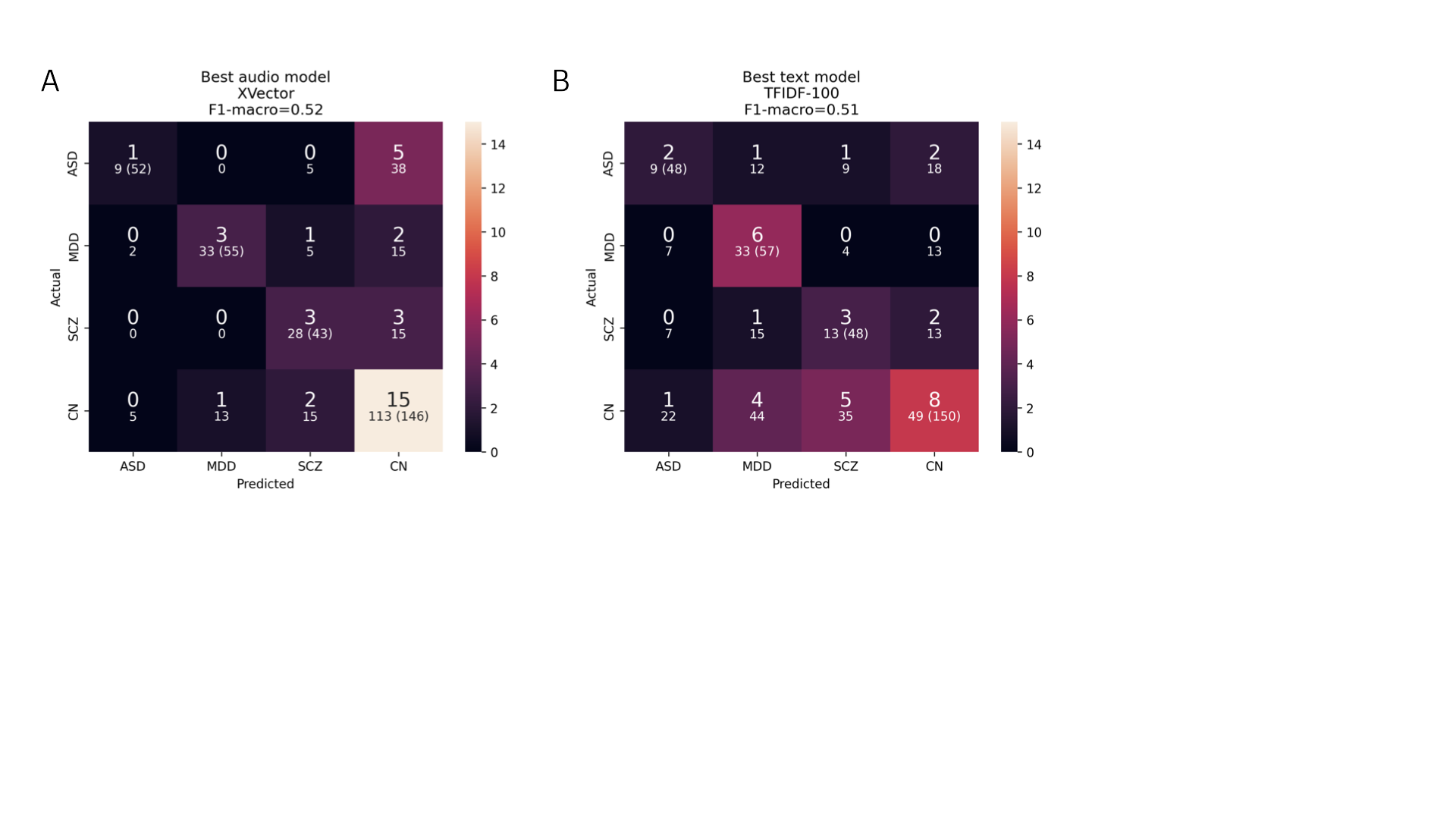}
    \caption{Confusion matrices of the performance of the best multiclass models on the test set. The large number indicates the number of participants identified to belong to the specific cell. The small number shows the number of trials (that is, recordings or transcripts) falling into the cell, with the number in parenthesis showing the total number of positive trials. For instance, in Panel A, 33 trials (recordings) were correctly classified as being from individuals with depression out of a total 55 recordings from individuals with depression. A: X-Vector. B: TF-IDF-100.}
    \label{fig:conf_mat}
\end{figure*}

As shown in the confusion matrices in \autoref{fig:conf_mat}, the error patterns for the best models from each modality were slightly different. On the test set, the best voice model (\textit{X-Vector}) over-predicted the healthy control group. The best text model (\textit{100-dimensional TF-IDF}) tended to over-diagnose MDD and schizophrenia, while only rarely predicting ASD. 
As shown in \autoref{fig:error_analysis}, age, gender and symptom severity did not seem to be systematically associated with misclassifications.

\section{Discussion} 
\label{sec:discussion}
    This study assessed the performance of machine learning models of voice- and text-based markers of neuropsychiatric conditions in a multi-diagnostic setting. We used a novel dataset including audio recordings from individuals with schizophrenia, MDD, and ASD, as well as matched controls for each group. We trained a variety of machine learning models on both binary classification (specific disorder vs. matched controls) and multiclass classification (identifying the correct diagnosis) to assess performance in problem settings that are closer to naturalistic clinical assessment. To comprehensively evaluate the potential of state-of-the-art speech-based models, we trained and evaluated models leveraging different input modalities (voice-based models, text-based models, and cross-modal ensembles), as well as architectures relying on both traditional feature-based approaches and state-of-the-art deep learning models (i.e., Transformers), which are designed to better capture contextual dependencies and the temporal dynamics of the input.  

\subsection{Binary vs. multi-diagnostic performance of voice and text models}
\label{sub:binary}

Our findings indicate that, generally, voice and transcripts contain information relevant to identifying patients \textit{vs} matched controls in binary settings. In line with previous work leveraging clean contrasts between patients and non-psychiatric controls, we were able to train binary models that performed better than highly conservative dummy baselines. The only exception was ASD, the smallest diagnostic group, where the best model performed only slightly better than our baseline. However, performance dropped for models trained in a multi-class context, where models are asked to discriminate between multiple diagnostic groups and against a more varied group of controls. The only notable exception to this was MDD, where performance was still fairly high. 

These findings suggest that, in binary classification settings, models may not be learning \textit{diagnosis-specific} and \textit{generalizable} markers of neuropsychiatric conditions, but rather develop heuristics that: a) apply to \textit{multiple} conditions; and/or b) only apply to samples drawn from specific matched socio-demographic groups. 

Concerning the first point, high performance in binary classification settings may be observed if models pick up on auditory and linguistic patterns that are found in \textit{multiple clinical conditions}, or if the model identifies markers related to clinical features that are present in multiple groups. To provide an example of the first scenario, both MDD and schizophrenia present decreased pitch variability in relation to negative and depressive symptoms. This is a useful feature to learn in a binary classification setting, but it does not help classify diagnoses in a multiclass setting that includes both groups. Analogously, anxiety is a common co-morbidity of many psychiatric conditions, including MDD, schizophrenia, and ASD; and has been shown to be detectable from speech \parencite{harrigan_role_1994}. 

Secondly, voice or text features that might seem relevant to discriminating between patients and controls in matched samples might be confounded with markers of socio-demographic factors such as age, gender, and education, thus failing to classify diagnoses in more varied samples. For instance, slow speech, increased pauses and simpler syntax have been found in schizophrenia, but they also characterize older adults. As a consequence, models fitting to these markers might not perform well in multi-diagnostic settings with a more varied population. To develop better heuristics, models may need to learn more complex features that require significantly larger datasets.

Importantly, there are currently no large-scale multidiagnostic datasets that include the fine-grained information on patients' symptoms and clinical characteristics required to test these hypotheses. In our view, developing these resources should be a major focus for future research, as assessing the clinical utility of speech-based models of neuropsychiatric condition relies on thorough assessment of these scenarios. We will expand upon this point in the next sections.

\subsection{Model performance across modalities and model types}

Text models generally performed better than audio models on multi-class classification, with 7 out of 10 among the best-performing models being text models (see \autoref{fig:multiclass_top_10}). The only exception to this pattern is a classifier based on embeddings from an X-Vector model, pretrained on speaker verification \parencite{speechbrain_speechbrainspkrec-xvect-voxceleb_2022}.

An important finding from the present study is that voice and text-based features provide \textit{complementary} information, which can be combined to improve multi-diagnostic performance. Indeed, we found that a simple ensemble pooling predictions of the best-performing audio and text baseline models outperformed all other models. 

Contextual embeddings from Transformer models did not seem to yield performance gains compared to more traditional features. While generally well-performing, Transformers were rarely the top-performing models, possibly due to the size of current datasets not being sufficient to adequately fine-tune these large architectures (see \autoref{sec:limitations} for a more in-depth discussion).

\subsection{Limitations}
\label{sec:limitations}

Our findings provide valuable insight on the specificity and generalizability of current machine learning approaches to speech markers in psychiatry. However, in line with the general landscape of the field, the study presents some limitations which must be overcome before we can develop robust, generalizable, and clinically useful markers of neuropsychiatric conditions. First, the novel dataset we developed for the present study includes 3 diagnostic groups, 420 participants and more than 3000 recordings. This is larger and more heterogeneous than most analogous datasets, usually involving fewer than 100 participants and clean binary contrasts \parencite{low_automated_2020}. Nevertheless, our dataset does not capture the full complexity of psychiatric conditions and their overlaps. More diagnostic groups, and more participants would be required, especially in the light of the overfitting (large performance differences between training and unseen test sets) we observe for many of our models. Furthermore, our data lacks more fine-grained meta-data to better characterize clinical heterogeneity and overlaps between diagnostic groups, for instance, cognitive abilities, anxiety and depression scores for all participants, and test the scenarios outlined in \autoref{sub:binary}.

Limitations in size and heterogeneity of the training data are likely to especially penalize Transformer-based models. Transformer models are significantly larger than feature-based models, and adequate fine-tuning of the full architecture requires more data than those required to train feature-based classifiers. This may be the reason we only observe limited gains compared to feature-based approaches. As we will outline in the next section, larger and more heterogeneous datasets, and domain-specific pretrained models may help address these limitations.

Approximately a quarter (26.4\%) of the individuals with schizophrenia included in this study were receiving treatment with antipsychotic medications, which may affect speech \parencite{thompson_pharmacological_1995}. This might make it harder for models to learn to identify vocal patterns of schizophrenia. 

Finally, in line with existing work, we focus on training models that predict diagnoses. But, diagnoses are not monolithic syndromes, and the "ontology" of psychiatric diagnoses itself is very much an open question \parencite{fried_studying_2021}. Individuals with the same diagnosis may exhibit radically different clinical profiles, and, conversely, different diagnoses may have substantial overlap in terms of symptoms and cognitive profiles, and comorbidities are far from infrequent. Different prediction targets might be preferable.

\subsection{Future directions}

In light of these findings and of the above-mentioned limitations, we advance a series of suggestions for future directions.

\subsubsection{Larger and more heterogeneous datasets}
Voice- and text-based models trained on binary classification of psychiatric diagnoses may lack the ability to learn specific and generalizable markers of psychiatric conditions, which limits their robustness and potential for real-world deployment. Machine learning research in this domain should therefore strive to tackle predictive problems that more adequately capture the complexity, heterogeneity and overlap of neuropsychiatric conditions, and that better mirror naturalistic application scenarios. 

At the moment, however, the lack of large datasets including data from multiple diagnostic groups is a major obstacle to this endeavour. Developing datasets that include speech samples from multiple diagnostic groups and report fine-grained clinical information for individual participants is a crucial step towards more robust speech-based psychiatric assessment.

Availability of larger and more heterogeneous datasets would also make it possible to experiment with more sophisticated architectures. In this study, for example, our models are trained and evaluated on generating separate diagnostic predictions for each recording. With more data, models could be trained on aggregated outputs from multiple recordings, allowing them to learn dependencies between participants' linguistic behavior across recordings and to formulate more informed predictions. Furthermore, here we have trained the simplest possible ensemble models by pooling estimates from audio and text models. Larger amounts of data would allow us to implement more sophisticated cross-modal pooling strategies, such as training classifiers on concatenated voice-and-text representations, or training ensemble Transformers.

\subsubsection{Better domain-specific models}
Against our expectations, pretrained text and audio Transformers did not outperform baseline models. Beyond the limited sample size available, an additional possible explanation for this is that encodings from models pre-trained on generic language modeling tasks, such as next-token (\textit{forward language modeling}) and missing token (\textit{masked language modeling}) prediction may not encode information suited to inferring psychiatric traits. Further fine-tuning of Transformer models on non-clinical data using self-supervised tasks \parencite{rocca_language_2022}, or supervised tasks (e.g., prediction of psychological traits) closer to the target domain, may yield performance improvements when pre-trained models are deployed for psychiatric applications.

\subsubsection{Better targets of prediction}
As anticipated in \autoref{sec:limitations}, a limitation of current work on speech markers of neuropsychiatric condition is its almost exclusive focus on predicting \textit{diagnoses} from voice and text. As diagnoses are heterogeneous and often overlapping, shifting the focus from predicting overall diagnostic groups towards predicting more fine-grained clinically relevant features (e.g. affect, cognitive load, etc.) could be a more robust and clinically useful endeavor \parencite{insel_research_2010}. Clinical assessment overwhelmingly relies on thorough testing of a range of such intermediate descriptors - more than of final outcomes such as diagnosis -  and machine learning models able to complement and support this assessment could make a substantial contribution and be of more immediate use. 

Being able to predict fine-grained clinical features could also guide practitioners beyond assessment, allowing them to monitor time-varying symptoms, to select between alternative treatments, or provide them with prognostic priors on, for instance, the probability of relapse. If predicting diagnoses is a worthy first step for an emerging field (with limited data), the largest impact will be given by focusing on predictive targets that can support and scale, rather than simply replicate, human decision-making.

\section{Conclusion} 
\label{sec:conclusion}
    We investigated the performance of a wide range of speech-based machine learning models of neuropsychiatric conditions in multi-class classification settings. We assessed the performance of voice-, transcript-based model, and cross-modal ensemble models, as well as the performance of traditional feature-based machine learning approaches and advanced state-of-the-art Transformer models. Models contrasting one specific diagnostic group with matched non-psychiatric controls performed largely in line with the literature. However, in multi-class contexts that better mirror naturalistic differential diagnosis, performance decreased quite drastically. Voice and text-based features complemented each other. Transformer models did not generally outperform baseline models. In order to develop models that can capture more robust and clinically relevant markers, we advocate for the field to: 1) develop larger and more heterogeneous datasets, better reflecting the complexity of neuropsychiatric conditions; 2) develop domain-specific Transformer models for the task at hand; 3) re-focus predictions from diagnoses to more fine-grained clinically relevant features, such as symptoms and cognitive abilities.
\bigskip

\hrule
\bigskip

\textbf{Author statement}: LH: Conceptualization; Methodology; Data Curation; Software; Formal Analysis; Validation; Writing - Original Draft; Writing - Review and Editing; Visualization; Interpretation; RR: Conceptualization; Methodology; Data Curation; Software; Formal Analysis; Validation; Writing - Original Draft; Writing - Review and Editing; Visualization; Interpretation; AS: Investigation; Data Curation; Methodology; Writing - Review and Editing; AP: Writing - Review and Editing; Interpretation; VB: Data Curation; Investigation; Resources; Writing - Review and Editing; NL: Data Curation; Investigation; Resources; Writing - Review and Editing; Interpretation; DB: Data Curation; Investigation; Resources; Writing - Review and Editing; Interpretation; KT: Investigation; Software; Writing - Review and Editing; Interpretation; EW: Investigation; Software; Writing - Review and Editing; Interpretation; SDØ: Writing - Review and Editing; Interpretation; RF: Conceptualization; Methodology; Data Curation; Formal Analysis; Validation; Writing - Original Draft; Writing - Review and Editing; Visualization; Supervision; Funding Acquisition; Project Administration; Interpretation

\bigskip

\textbf{Acknowledgements}: We acknowledge seed funding from the Interacting Minds Centre ("Clinical voices"). RR is partly supported by funding from the Volkswagen Stiftung. DB is supported by a Sir Henry Wellcome Postdoctoral Fellowship from the Wellcome Trust (213630/Z/18/Z). The Wellcome Centre for Human Neuroimaging is supported by core funding from the Wellcome Trust (203147/Z/16/Z). 

\bigskip

\textbf{Conflicts of interest}
SDØ received the 2020 Lundbeck Foundation Young Investigator Prize. Furthermore, SDØ owns/has owned units of mutual funds with stock tickers DKIGI, IAIMWC, SPIC25KL and WEKAFKI, and has owned units of exchange traded funds with stock tickers BATE, TRET, QDV5, QDVH, QDVE, SADM, IQQH, USPY, EXH2, 2B76 and EUNL.

\bigskip
\hrule
\smallskip

\printbibliography
\renewcommand{\glsgroupskip}{}
\printglossary[nonumberlist]

\appendix
\counterwithin{figure}{section}
\counterwithin{table}{section}
\label{sec:appendix}

\label{sec:appendix:original_studies}
\section{Original studies}

The depression corpus was collected by \textcite{ladegaard_higher-order_2014} for a study on changes in social cognition in first-episode depression. The corpus consists of 42 patients with first-episode \gls{mdd}, 42 pairwise matched controls on age, sex, and educational background, and 25 patients with chronic depression \parencite{ladegaard_course_2016}. Patients were recruited from general practitioners in the Central Denmark Region, and healthy controls were recruited via newspaper advertisements and offered monetary compensation if included in the study (1000 DKK). All participants were native speakers of Danish and met the following inclusion criteria: 1) first-episode major depression was the primary diagnosis, 2) the severity of depression was moderate to severe as measured by the 17-item Hamilton Rating Scale for Depression (HamD-17 > 17), 3) patients were psychotropic drug-naïve, 4) no psychiatric comorbidity other than anxiety disorders. Patients with head trauma, neurological illness, or substance use disorders were not permitted to the study. Exclusion criteria for healthy controls were the same as for depressed patients.

The schizophrenia corpus comes from \textcite{bliksted_effect_2017, veddum_patients_2019} and consists of recordings from 104 patients with schizophrenia and 116 healthy controls pairwise matched on age, sex, level of education, handedness, race, community of residency, parental social economic status, and expected parental income. Patients were recruited from the OPUS Clinic for young people with schizophrenia and were diagnosed with first-episode schizophrenia according to ICD-10 and received less than 3 months of antipsychotic medication prior to the diagnostic interview. Exclusion criteria included history of neurological disorder or severe head trauma, ICD-10 diagnosis of drug- or alcohol dependency, estimated premorbid IQ < 70, and not being able to understand spoken Danish sufficiently to comprehend the testing procedures. Healthy controls were recruited via newspaper advertisements. 

\begin{table*}[!ht]
\centering
\caption{Descriptive statistics of the audio data. Rows are paired to indicate the patients and corresponding matched controls. }
\label{app_tab:audio_description}
\begin{tabularx}{\textwidth}{lrrrrr}
\toprule
Diagnosis & N participants & N recordings &  Mean duration (s) &  Total duration (h) \\
\midrule
ASD &             46 &          424 &               40.0 &                 4.7 \\
CN &             41 &          382 &               29.1 &                 3.1 \\
\cmidrule{1-1}
MDD &             66 &          548 &               13.3 &                 2.0 \\
CN &             41 &          372 &               12.7 &                 1.3 \\
\cmidrule{1-1}
SCZ &            105 &          821 &               19.2 &                 4.4 \\
CN &            117 &          969 &               19.1 &                 5.2 \\
\midrule
Total &           416 &         3516 &               21.2 &                20.7 \\
\bottomrule
\end{tabularx}
\end{table*}
\begin{table*}[!ht]
\centering
\caption{Descriptive statistics of the transcriptions data. Rows are paired to indicate the patients and corresponding matched controls.}
\label{app_tab:transcript_description}
\begin{tabularx}{\textwidth}{lrrrr}
\toprule
Diagnosis & N participants & N transcripts &  Mean number of words &  Total number of words \\
\midrule
ASD &             21 &           162 &             65.9 &                  10,691\\
CN &             25 &           200 &             73.5 &                  14,711\\
\cmidrule{1-1}
MDD &             67 &           585 &             28.9 &                  16,959 \\
CN &             42 &           408 &             34.6 &                  14,124 \\
\cmidrule{1-1}
SCZ &            105 &           873 &             43.4 &                  37,917  \\
CN &            116 &           990 &             48.5 &                  48,112 \\
\midrule
Total&            376 &          3218 &             44.2 &                 142,514 \\
\bottomrule
\end{tabularx}
\end{table*}

The \gls{asd} corpus consists of 46 patients, and 41 pairwise matched controls and was not otherwise published before this study. Patients were included based on a self-reported diagnosis of autism spectrum disorder. 

\label{sec:appendix:triangles}
\section{Animated Triangles Task}
The Animated Triangles task is commonly used to measure theory of mind and involves eight to twelve video clips of interactions between animated triangles. In four of the clips, the triangles are moving randomly and unintentionally (e.g. bouncing off walls). In the remaining clips, the triangles are interacting intentionally; in four of them, the triangles are trying to influence the mental state of one another (e.g., a larger triangle trying to convince a small one to leave a closure), and in the remaining four the triangles are performing an activity together. Each animation has a duration of  approximately 40 seconds, after which participants are asked to provide an interpretation of what was happening in the animation. This interpretation was recorded and forms the basis for this study. 

\label{sec:appendix:preprocessing}
\section{Data Processing}
\subsection{Audio data preprocessing}
All recordings were first manually processed to identify issues with sound quality, and remove long periods of silence and interviewer speech. Subsequently, background noise was removed using iZotope Elements RX6 \parencite{izotope_rx_2017}. This was done to reduce the chance of models fitting to group-specific patterns of background noise. Recordings containing less than 3 seconds of speech were removed, as they are unlikely to contain enough signal to be useful. Recordings were resampled to 16kHz as that is required by models such as Wav2vec2 \parencite{baevski_wav2vec_2020} and XLS-R \parencite{babu_xls-r_2021}, and audio samples were normalized to lie between -1.0 and 1.0. 

Before training, all recordings were split into 5 second segments with a 1 second striding window. That is, a recording with a duration of 7 seconds was split into three separate recordings: 0-5 seconds, 1-6 seconds, and 2-7 seconds. Windowing was done to increase the amount of training data, and serves as a form of augmentation. Additionally, it is computationally costly to process audio, so short windows allowed us to use a larger batch size. See \autoref{fig:pipeline}(B) for a visualization.

\subsection{Transcript data preprocessing}
All audio files with sufficient audio quality were manually transcribed by research assistants. Transcripts were manually checked for errors and consistency. For instance, some transcribers, due to software issues, transcribed æ, ø, å as ae, oe, and aa, respectively, which was standardized. Analogously, transcribers included comments in square parentheses (e.g. '[sic]'), which were also excluded.

\label{sec:appendix:models}
\section{Models}
The following two subsections give a detailed explanation of the audio and text models. For each modality, multiple types of models were created; baselines and Transformer models. Each model type was trained to predict the correct disorder in a multiclass setting, as well as separate binary models that were trained to discriminate each disorder from their matching controls. 

\subsection{Audio Models}
The following section describes the development of baseline and Transformer models for audio and outlines the training procedure, hyperparameters, and augmentation strategy. The training setup and procedure were the same across all audio models to aid comparability: the only varying factor was the features used.

\subsubsection{Baseline Models}
For consistency with previous studies of inferring mental disorder from voice, two baseline models were created using the eGeMAPS and ComParE feature sets extracted using the \textit{openSMILE} Python library v2.4.1 \parencite{eyben_opensmile::_2015}. A third baseline model was created using 40 MFCCs extracted with \textit{torchaudio} v0.11.0. A final baseline model was trained using features extracted from an X-vector model \parencite{snyder_x-vectors_2018} pretrained on the multilingual VoxCeleb1 \parencite{nagrani_voxceleb_2017} and VoxCeleb2 \parencite{chung_voxceleb2_2018} datasets for speaker recognition \parencite{speechbrain_speechbrainspkrec-xvect-voxceleb_2022}. The X-vector model is based on temporal/1D convolutional neural networks (CNN) to utilize both temporal and acoustic information from speech. We include the X-vector model to provide a comparison to a different type of pretrained model that is CNN instead of Transformer based. 

The extracted baseline features were aggregated over each audio window using the mean, to obtain a single feature embedding for each window. The number of features extracted from each baseline feature method is shown in \autoref{multidiag:tab:n_features}.

\begin{table}[]
\centering
\caption{Number of features per feature set}
\label{multidiag:tab:n_features}
\begin{tabular}{lr}
\toprule
Feature set & N. features \\ \midrule
MFCCs       & 40          \\ 
eGeMAPS     & 88          \\
X-Vector    & 512         \\
Wav2vec2    & 1024       \\
ComParE     & 6377       \\
\bottomrule
\end{tabular}
\end{table}

\subsubsection{Transformer Models}
At present there are three pretrained Transformer models for Danish speech publicly available; a Danish Wav2vec2 base model trained by Alvenir, \textit{wav2vec2-base-da}, \parencite{alvenir_alvenirwav2vec2-base-da_2022}, the multilingual \textit{XLS-R} trained by Facebook \parencite{babu_xls-r_2021}, and a Danish continued pretraining of \textit{XLS-R} \parencite{hansen_xls-r-300m-danish_2022}. \textit{wav2vec2-base-da} has approximately 95 million trainable parameters and was pretrained on 1,300 hours of Danish speech, mainly from audiobooks. \textit{XLS-R} was pretrained on a total of 436 thousand hours of speech from 53 different languages and has approximately 300 million trainable parameters. Danish makes up approximately 14 thousand hours of the training data, and mainly stems from speeches held in the European Parliament in Danish. The Danish \textit{XLS-R} warmstarted from the \textit{XLS-R} checkpoint and further pretrained on 141,000 hours of Danish radio.

We use all three models as feature extractors by freezing the weights of the entire model and training a single classification layer that is added on top. As with the baseline models, we aggregate the extracted features over each recording using the mean before feeding them to the classification layer. 

\subsubsection{Model Training}\label{multidiag:sec:augmentation}
Audio models were trained using the \textit{PyTorch Lightning} v1.6.3 \parencite{falcon_pytorch_2022} and \textit{transformers} v4.19.2 \parencite{wolf_transformers_2020} Python libraries on a single NVIDIA Tesla V100 32GB GPU. For both baselines and transformers, model training progressed in the following steps:

\begin{enumerate}
    \item Load batch of windowed data. 
    \item Randomly augment the data to increase or decrease gain, add coloured noise (brown, white, pink, or blue), add an impulse response, and/or use \textit{SpliceOut}. 
    \item Extract features from the augmented data using either a baseline feature extractor or a Transformer model.
    \item Pass features to a shallow neural network consisting of \textit{n\_features} input units, a hidden layer with 1024 units, followed by an output layer with 2 or 4 units for the binary and multiclass task respectively. 
    \item Update weights of the shallow neural network using the AdamW optimizer. \parencite{loshchilov_decoupled_2019}. If multiclass, loss was weighted by the number of examples from each class to reduce the chance of over-predicting healthy controls.
    \item Iterate for 20 epochs.
\end{enumerate}

The batch size was set to 32 and models were trained to minimize cross-entropy. The learning rate was set to $0.00005$ for all models after initial experimentation.

Augmentation was done using \textit{torch\_audiomentations} v0.10.1 \parencite{jordal_asteroid-teamtorch-audiomentations_2022}. The idea of data augmentation is to inject noise or change the input in subtle ways to expose the models to more diverse input representations. Data augmentation has been successfully used in both \gls{cv} \parencite{perez_effectiveness_2017}, \gls{nlp} \parencite{wei_eda_2019}, and speech analysis \parencite{park_specaugment_2019}, to create more robust and generalizable models.

On each iteration, each example had a 50\% chance of being augmented by each of the augmentations described above. The gain augmentation changes the gain of the recording by a number of decibels uniformly sampled between -18 and 6. Coloured noise adds noise with peaks at different frequencies depending on colour. The impulse responses were acquired from \textcite{traer_statistics_2016}, and are a collection of 271 impulse responses from places encountered by 7 participants during their daily life. An impulse response is the reverberation profile of a location and can vary greatly. For instance, the reverberation in a church is very different from in a forest. SpliceOut was introduced by \textcite{jain_spliceout_2021} and is a computationally more efficient variation of SpecAugment \parencite{park_specaugment_2019}, which masks random time intervals of the input audio.
The reason for training a shallow neural classifier on the feature sets instead of an XGBoost model (as for text), was to enable the use of varying augmentations per epoch.

\subsection{Text Models}
The following section describes the development of the text models and outlines the training procedure, hyperparameters, and feature sets. Baseline models were trained using XGBoost, as the performance was found to be poor using a shallow neural network in a similar fashion as for the audio models and text Transformers. 

\subsubsection{Baseline Models}
Multiple types of baseline models were constructed. One type using statistical descriptors of text, extracted using the \textit{TextDescriptives} v.1.0.7 Python library \parencite{hansen_textdescriptives_2023}, one using \gls{tfidf} features, and one using pretrained \textit{fasttext} word embeddings \parencite{bojanowski_enriching_2016}. \autoref{tab:text_features} 
displays the different features used. Separate models were trained for each feature, as well as a model using all the descriptive, readability, syntactic complexity, and sentiment features.

\subsubsection{Transformer Models}
We trained four publicly available Transformer models for Danish: \textit{ScandiBERT} \parencite{snaebjarnarson_vesteinnscandibert_2022}, \textit{XLM-RoBERTa-base} \parencite{conneau_unsupervised_2020}, \textit{Multilingual MiniLM Sentence Transformer} \parencite{reimers_making_2020}, and \textit{Danish Electra} \parencite{hojmark-bertelsen_aelaectra_2021}. \textit{ScandiBERT} was pretrained on Danish, Norwegian, and Swedish text, \textit{XLM-RoBERTa-base} and \textit{Multilingual MiniLM Sentence Transformer} were pretrained for 100 different languages, and \textit{Danish Electra} was trained only on Danish text.

\begin{table}[H]

\caption{Features used for text baselines.}
\label{tab:text_features}
\resizebox{\columnwidth}{!}{
\begin{tabular}[t]{l}
\toprule
\addlinespace[0.3em]
\multicolumn{1}{l}{\textbf{Descriptive}}\\
\midrule
\hspace{1em}Sentence length mean and std\\
\hspace{1em}Token length mean and std\\
\hspace{1em}Proportion unique tokens\\
\hspace{1em}Number of unique tokens\\
\hspace{1em}Number of characters\\
\hspace{1em}Number of sentences\\
\addlinespace[0.3em]
\multicolumn{1}{l}{\textbf{Readability}}\\
\midrule
\hspace{1em}Flesch reading ease\\
\hspace{1em}Flesch-Kincaid grade\\
\hspace{1em}Lix\\
\hspace{1em}Rix\\
\hspace{1em}Coleman-Liau Index\\
\addlinespace[0.3em]
\multicolumn{1}{l}{\textbf{Syntactic complexity}}\\
\midrule
\hspace{1em}Dependency distance mean and std\\
\hspace{1em}Proportion adjacent dependency relations mean and std\\
\addlinespace[0.3em]
\multicolumn{1}{l}{\textbf{Sentiment}}\\
\midrule
\hspace{1em}Polarity\\
\addlinespace[0.3em]
\multicolumn{1}{l}{\textbf{TF-IDF}}\\
\midrule
\hspace{1em}100 features\\
\hspace{1em}1.000 features\\
\hspace{1em}10.000 features\\
\addlinespace[0.3em]
\multicolumn{1}{l}{\textbf{Fasttext}}\\
\midrule
\hspace{1em}Danish 300-dimensional Fasttext word vectors\\
\bottomrule
\end{tabular}
}
\end{table}

\subsubsection{Model Training}
Baseline models were trained using \textit{scikit-learn} v1.1.1 \parencite{pedregosa_scikit-learn:_2011}, and Transformers using \textit{transformers} v4.19.2 \parencite{wolf_transformers_2020} and \textit{PyTorch Lightning} v1.6.3 \parencite{falcon_pytorch_2022}. XGBoost hyperparameters for baseline models were found via a grid search of the parameters in \autoref{tab:xgboost_grid} and optimized on the validation set. 

\begin{table}[H]
\centering
\caption{Grid for XGBoost hyperparameter search.}
\label{tab:xgboost_grid}
\resizebox{\columnwidth}{!}{
\begin{tabular}{ll}
\toprule
Parameter          & Values                      \\ 
\midrule
\texttt{learning\_rate}     & {[}0.001, 0.01, 0.1, 0.5{]} \\
\texttt{min\_child\_weight} & {[}1, 3, 5, 10{]}           \\
\texttt{gamma}              & {[}0, 0.5, 1, 2{]}          \\
\texttt{subsample}          & {[}0.6, 0.8, 1{]}           \\
\texttt{colsample\_bytree}  & {[}0.25, 0.5, 0.75, 1{]}    \\
\texttt{max\_depth}         & {[}1, 3, 5{]}               \\
\texttt{reg\_alpha}         & {[}0, 0.1, 1, 5{]}          \\
\texttt{reg\_lambda}        & {[}0.1, 1, 5{]}             \\
\texttt{n\_estimators}      & {[}20, 100, 500{]}          \\ 
\bottomrule
\end{tabular}
}
\end{table}

Text transformers were trained by adding a classification layer on top of the base Transformer architectures in a similar manner as for the audio models (i.e. a linear layer feeding to an output layer with 2 or 4 outputs for the binary and multiclass task, respectively). Transformer models were trained both with a frozen base model, i.e. by only updating the parameters of the newly initialized classification layer, and by updating the entire Transformer. The learning rate for the Transformer models was searched for among values of 1 to the negative power of 2 to 6, and optimized on the validation set. After parameter optimization, all Transformer models were trained with a batch size of 16 for 100 epochs. 

Similarly to audio models, the loss of all text models was weighted by the number of examples from each class.

\subsection{Multi-modal Combination}
To estimate whether additional performance could be gained by combining text and audio models, we combined predictions by choosing the output of the most confident model. Two ensembles were created for each task: one created from the best performing baseline model from each modality, and one from the best performing Transformer model from each modality. The best-performing models were chosen based on F1-macro on the test set. This was done to eliminate the need to create all possible combinations of models, but rather to see if the combination of two good models yielded an improvement. 

\section{Performances}
Tables and figures reporting performance metrics for all models, as well as details on our error analysis, are displayed in the next pages.
\begin{table*}
\setlength{\tabcolsep}{2.8pt}
\caption{Performances for all multiclass models}
\label{tab:multiclass_perf}
\centering
\resizebox{0.85\textwidth}{!}{
\begin{adjustbox}{height=9cm, center}
\begin{tabular}
{rc||ccc||ccc||ccc||ccc}
\toprule
\multicolumn{1}{c}{} & \multicolumn{1}{c}{} & \multicolumn{3}{c}{F1 Macro} & \multicolumn{3}{c}{F1 - ASD} & \multicolumn{3}{c}{F1 - MDD} & \multicolumn{3}{c}{F1 - SCZ} \\
                                        &&               Test & Validation & Train &     Test & Validation & Train &      Test & Validation & Train &      Test & Validation & Train \\
\midrule
                      Baseline Ensemble & ensemble &              0.570 &      0.476 & 0.814 &    0.250 &      0.286 & 0.845 &     0.667 &      0.400 & 0.829 &     0.667 &      0.571 & 0.764 \\
                               X-Vector &    audio &              0.521 &      0.497 & 0.789 &    0.286 &      0.286 & 0.812 &     0.600 &      0.444 & 0.777 &     0.500 &      0.571 & 0.756 \\
                          TF-IDF, 100-d &     text &              0.511 &      0.432 & 0.642 &    0.444 &      0.167 & 0.615 &     0.667 &      0.500 & 0.654 &     0.400 &      0.500 & 0.630 \\
                   Transformer Ensemble & ensemble &              0.461 &      0.274 & 0.811 &    0.286 &      0.214 & 0.773 &     0.667 &      0.500 & 0.894 &     0.375 &      0.286 & 0.828 \\
                   Sentence Transformer &     text &              0.461 &      0.455 & 0.869 &    0.286 &      0.286 & 0.947 &     0.667 &      0.588 & 0.894 &     0.375 &      0.333 & 0.832 \\
                            XLM-RoBERTa &     text &              0.452 &      0.499 & 0.643 &    0.444 &      0.400 & 0.545 &     0.750 &      0.462 & 0.736 &     0.000 &      0.500 & 0.575 \\
                               FastText &     text &              0.450 &      0.354 & 0.597 &    0.400 &      0.333 & 0.545 &     0.625 &      0.444 & 0.687 &     0.375 &      0.200 & 0.648 \\
                             ScandiBERT &     text &              0.444 &      0.450 & 0.698 &    0.444 &      0.400 & 0.571 &     0.667 &      0.533 & 0.780 &     0.167 &      0.250 & 0.701 \\
                                Ælectra &     text &              0.413 &      0.421 & 0.607 &    0.444 &      0.400 & 0.545 &     0.571 &      0.545 & 0.675 &     0.154 &      0.222 & 0.620 \\
                       All descriptives &     text &              0.377 &      0.425 & 0.613 &    0.250 &      0.250 & 0.667 &     0.300 &      0.500 & 0.560 &     0.444 &      0.400 & 0.522 \\
                         TF-IDF, 1000-d &     text &              0.375 &      0.374 & 0.592 &    0.364 &      0.182 & 0.552 &     0.632 &      0.400 & 0.645 &     0.133 &      0.400 & 0.604 \\
                                  MFCCs &    audio &              0.373 &      0.337 & 0.669 &    0.167 &      0.250 & 0.644 &     0.462 &      0.182 & 0.680 &     0.250 &      0.375 & 0.627 \\
Proportion Adjacent Dependencies (mean) &     text &              0.368 &      0.305 & 0.455 &    0.222 &      0.182 & 0.409 &     0.250 &      0.222 & 0.486 &     0.364 &      0.167 & 0.262 \\
                   Number of characters &     text &              0.334 &      0.273 & 0.385 &    0.250 &      0.167 & 0.378 &     0.476 &      0.364 & 0.472 &     0.000 &      0.000 & 0.082 \\
                    Flesch reading ease &     text &              0.333 &      0.410 & 0.461 &    0.250 &      0.167 & 0.400 &     0.182 &      0.588 & 0.408 &     0.286 &      0.400 & 0.411 \\
                Proportion unique okens &     text &              0.316 &      0.169 & 0.543 &    0.000 &      0.200 & 0.560 &     0.364 &      0.100 & 0.505 &     0.400 &      0.000 & 0.475 \\
                                    RIX &     text &              0.316 &      0.329 & 0.306 &    0.000 &      0.000 & 0.154 &     0.429 &      0.429 & 0.421 &     0.286 &      0.333 & 0.040 \\
                 Sentence length (mean) &     text &              0.313 &      0.369 & 0.576 &    0.000 &      0.000 & 0.667 &     0.400 &      0.222 & 0.497 &     0.267 &      0.588 & 0.440 \\
              Syllables per token (std) &     text &              0.303 &      0.309 & 0.526 &    0.333 &      0.267 & 0.360 &     0.364 &      0.400 & 0.573 &     0.000 &      0.235 & 0.479 \\
                     Token length (std) &     text &              0.289 &      0.312 & 0.270 &    0.182 &      0.182 & 0.092 &     0.500 &      0.471 & 0.397 &     0.000 &      0.000 & 0.040 \\
             Dependency distance (mean) &     text &              0.289 &      0.261 & 0.618 &    0.286 &      0.000 & 0.485 &     0.000 &      0.286 & 0.660 &     0.222 &      0.333 & 0.608 \\
 Proportion Adjacent Dependencies (std) &     text &              0.286 &      0.277 & 0.324 &    0.182 &      0.375 & 0.203 &     0.400 &      0.000 & 0.419 &     0.000 &      0.250 & 0.216 \\
                                Alvenir &    audio &              0.281 &      0.116 & 0.216 &    0.375 &      0.263 & 0.211 &     0.462 &      0.200 & 0.560 &     0.000 &      0.000 & 0.000 \\
                       Number of tokens &     text &              0.276 &      0.286 & 0.243 &    0.353 &      0.375 & 0.111 &     0.381 &      0.286 & 0.465 &     0.000 &      0.000 & 0.021 \\
                   Flesch-Kincaid grade &     text &              0.273 &      0.312 & 0.394 &    0.118 &      0.154 & 0.203 &     0.182 &      0.353 & 0.416 &     0.222 &      0.286 & 0.315 \\
                      Gunning Fog index &     text &              0.265 &      0.343 & 0.408 &    0.154 &      0.000 & 0.296 &     0.167 &      0.471 & 0.434 &     0.200 &      0.286 & 0.277 \\
                   Number unique tokens &     text &              0.253 &      0.246 & 0.402 &    0.182 &      0.118 & 0.412 &     0.300 &      0.273 & 0.477 &     0.000 &      0.286 & 0.100 \\
                        TF-IDF, 10000-d &     text &              0.251 &      0.374 & 0.474 &    0.000 &      0.167 & 0.387 &     0.571 &      0.400 & 0.560 &     0.125 &      0.500 & 0.490 \\
                                    Lix &     text &              0.229 &      0.332 & 0.451 &    0.200 &      0.200 & 0.391 &     0.167 &      0.375 & 0.463 &     0.000 &      0.182 & 0.286 \\
             Syllables per token (mean) &     text &              0.228 &      0.372 & 0.277 &    0.348 &      0.333 & 0.103 &     0.000 &      0.286 & 0.275 &     0.000 &      0.222 & 0.237 \\
            Automated readability index &     text &              0.219 &      0.263 & 0.658 &    0.000 &      0.000 & 0.593 &     0.182 &      0.364 & 0.636 &     0.143 &      0.125 & 0.667 \\
                       Number sentences &     text &              0.217 &      0.174 & 0.154 &    0.375 &      0.273 & 0.088 &     0.387 &      0.320 & 0.397 &     0.000 &      0.000 & 0.021 \\
                                eGeMAPS &    audio &              0.212 &      0.278 & 0.374 &    0.182 &      0.200 & 0.475 &     0.000 &      0.286 & 0.389 &     0.000 &      0.000 & 0.000 \\
                     Coleman-Liau index &     text &              0.203 &      0.404 & 0.587 &    0.000 &      0.250 & 0.545 &     0.154 &      0.471 & 0.516 &     0.143 &      0.308 & 0.575 \\
                      Mean token length &     text &              0.202 &      0.214 & 0.401 &    0.267 &      0.125 & 0.232 &     0.000 &      0.143 & 0.437 &     0.000 &      0.000 & 0.317 \\
                                  XLS-R &    audio &              0.202 &      0.189 & 0.085 &    0.364 &      0.312 & 0.191 &     0.000 &      0.000 & 0.000 &     0.000 &      0.000 & 0.021 \\
                      Sentence polarity &     text &              0.187 &      0.175 & 0.171 &    0.261 &      0.200 & 0.105 &     0.000 &      0.000 & 0.000 &     0.000 &      0.000 & 0.000 \\
                                  Dummy & baseline &              0.167 &      0.167 & 0.163 &    0.000 &      0.000 & 0.000 &     0.167 &      0.167 & 0.163 &     0.167 &      0.167 & 0.163 \\
                  Sentence length (std) &     text &              0.136 &      0.300 & 0.210 &    0.273 &      0.300 & 0.116 &     0.273 &      0.400 & 0.439 &     0.000 &      0.500 & 0.134 \\
              Dependency distance (std) &     text &              0.129 &      0.122 & 0.129 &    0.214 &      0.276 & 0.107 &     0.300 &      0.211 & 0.410 &     0.000 &      0.000 & 0.000 \\
                            Gjallarhorn &    audio &              0.116 &      0.097 & 0.097 &    0.263 &      0.222 & 0.155 &     0.000 &      0.000 & 0.000 &     0.200 &      0.167 & 0.235 \\
                                ComParE &    audio &              0.071 &      0.071 & 0.106 &    0.000 &      0.000 & 0.000 &     0.000 &      0.000 & 0.000 &     0.286 &      0.286 & 0.426 \\
\bottomrule
\end{tabular}
\end{adjustbox}
}
\end{table*}
\begin{table*}
\caption{Binary performances for MDD models}
\label{tab:depr_perf}
\begin{adjustbox}{height=10cm,center}
\setlength{\tabcolsep}{2.8pt}
\begin{tabular}{rc||ccc}
\toprule
\multicolumn{1}{c}{} & \multicolumn{1}{c}{} & \multicolumn{3}{c}{F1 - MDD} \\
                                        &&      Test & Train & Validation \\
\midrule
                                Ælectra &     text &     0.923 & 1.000 &      0.444 \\
                   Transformer Ensemble & ensemble &     0.923 & 1.000 &      0.444 \\
                             ScandiBERT &     text &     0.800 & 1.000 &      0.800 \\
                      Baseline Ensemble & ensemble &     0.714 & 0.713 &      0.667 \\
                            XLM-RoBERTa &     text &     0.714 & 0.800 &      0.400 \\
                   Sentence Transformer &     text &     0.714 & 1.000 &      0.600 \\
                  Sentence length (std) &     text &     0.714 & 0.688 &      0.571 \\
                                  MFCCs &    audio &     0.714 & 0.714 &      0.667 \\
                                eGeMAPS &    audio &     0.667 & 0.848 &      0.750 \\
                                ComParE &    audio &     0.667 & 0.788 &      0.667 \\
                                  XLS-R &    audio &     0.667 & 0.788 &      0.667 \\
                     Token length (std) &     text &     0.667 & 0.633 &      0.250 \\
                            Gjallarhorn &    audio &     0.667 & 0.779 &      0.667 \\
                                  Dummy & baseline &     0.667 & 0.786 &      0.667 \\
            Automated readability index &     text &     0.615 & 0.729 &      0.667 \\
             Dependency distance (mean) &     text &     0.615 & 0.718 &      0.500 \\
                          TF-IDF, 100-d &     text &     0.615 & 0.800 &      0.250 \\
                                    Lix &     text &     0.545 & 0.638 &      0.667 \\
                               X-Vector &    audio &     0.545 & 0.821 &      0.667 \\
                    Flesch reading ease &     text &     0.545 & 0.788 &      0.667 \\
                         TF-IDF, 1000-d &     text &     0.533 & 0.982 &      0.667 \\
                      Sentence polarity &     text &     0.533 & 0.696 &      0.714 \\
                        TF-IDF, 10000-d &     text &     0.500 & 0.472 &      0.286 \\
                                    RIX &     text &     0.500 & 0.713 &      0.615 \\
                   Flesch-Kincaid grade &     text &     0.500 & 0.745 &      0.615 \\
                   Number unique tokens &     text &     0.462 & 0.674 &      0.364 \\
 Proportion Adjacent Dependencies (std) &     text &     0.444 & 0.421 &      0.400 \\
              Syllables per token (std) &     text &     0.400 & 0.841 &      0.400 \\
                       All descriptives &     text &     0.400 & 0.739 &      0.667 \\
                 Sentence length (mean) &     text &     0.400 & 0.800 &      0.667 \\
                               FastText &     text &     0.400 & 0.972 &      0.727 \\
                                Alvenir &    audio &     0.364 & 0.617 &      0.286 \\
                       Number of tokens &     text &     0.364 & 0.755 &      0.600 \\
                      Gunning Fog index &     text &     0.364 & 0.747 &      0.667 \\
                     Coleman-Liau index &     text &     0.222 & 0.652 &      0.667 \\
Proportion Adjacent Dependencies (mean) &     text &     0.222 & 0.596 &      0.571 \\
                Proportion unique okens &     text &     0.200 & 0.622 &      0.400 \\
                      Mean token length &     text &     0.200 & 0.535 &      0.667 \\
             Syllables per token (mean) &     text &     0.000 & 0.164 &      0.000 \\
                       Number sentences &     text &     0.000 & 0.000 &      0.000 \\
                   Number of characters &     text &     0.000 & 0.568 &      0.286 \\
              Dependency distance (std) &     text &     0.000 & 0.000 &      0.000 \\
\bottomrule
\end{tabular}
\end{adjustbox}
\end{table*}

\begin{table*}
\caption{Binary performances for schizophrenia models}
\label{tab:schz_perf}
\begin{adjustbox}{height=10cm, center}
\setlength{\tabcolsep}{2.8pt}
\begin{tabular}{rc||ccc}
\toprule
\multicolumn{1}{c}{}  & \multicolumn{1}{c}{} & \multicolumn{3}{c}{F1 - SCZ} \\
                                        &&      Test & Validation & Train \\
\midrule
                      Baseline Ensemble & ensemble &     0.833 &      0.800 & 0.887 \\
                         TF-IDF, 1000-d &     text &     0.833 &      0.600 & 0.816 \\
              Dependency distance (std) &     text &     0.800 &      0.400 & 0.613 \\
                                Alvenir &    audio &     0.769 &      0.533 & 0.619 \\
                            Gjallarhorn &    audio &     0.769 &      0.462 & 0.554 \\
                            XLM-RoBERTa &     text &     0.750 &      0.667 & 0.984 \\
                             ScandiBERT &     text &     0.727 &      0.500 & 1.000 \\
                               X-Vector &    audio &     0.727 &      0.800 & 0.873 \\
                   Transformer Ensemble & ensemble &     0.727 &      0.500 & 0.989 \\
                       Number sentences &     text &     0.714 &      0.444 & 0.519 \\
                      Sentence polarity &     text &     0.667 &      0.667 & 0.641 \\
             Syllables per token (mean) &     text &     0.667 &      0.667 & 0.633 \\
                                  Dummy & baseline &     0.667 &      0.667 & 0.641 \\
             Dependency distance (mean) &     text &     0.667 &      0.545 & 0.568 \\
                                ComParE &    audio &     0.667 &      0.667 & 0.639 \\
                                  XLS-R &    audio &     0.667 &      0.667 & 0.639 \\
Proportion Adjacent Dependencies (mean) &     text &     0.615 &      0.667 & 0.652 \\
                       All descriptives &     text &     0.600 &      0.571 & 0.626 \\
                                  MFCCs &    audio &     0.600 &      0.833 & 0.798 \\
                                eGeMAPS &    audio &     0.571 &      0.429 & 0.654 \\
                  Sentence length (std) &     text &     0.571 &      0.615 & 0.625 \\
                        TF-IDF, 10000-d &     text &     0.545 &      0.727 & 0.897 \\
                Proportion unique okens &     text &     0.545 &      0.462 & 0.643 \\
                               FastText &     text &     0.545 &      0.667 & 0.750 \\
                 Sentence length (mean) &     text &     0.545 &      0.727 & 0.642 \\
                    Flesch reading ease &     text &     0.545 &      0.500 & 0.656 \\
              Syllables per token (std) &     text &     0.545 &      0.615 & 0.578 \\
                          TF-IDF, 100-d &     text &     0.545 &      0.444 & 0.660 \\
                                Ælectra &     text &     0.500 &      0.727 & 0.793 \\
            Automated readability index &     text &     0.500 &      0.769 & 0.705 \\
                   Number of characters &     text &     0.500 &      0.615 & 0.641 \\
                   Flesch-Kincaid grade &     text &     0.500 &      0.600 & 0.695 \\
                      Mean token length &     text &     0.462 &      0.545 & 0.641 \\
                   Number unique tokens &     text &     0.444 &      0.545 & 0.565 \\
                       Number of tokens &     text &     0.400 &      0.500 & 0.560 \\
                      Gunning Fog index &     text &     0.400 &      0.545 & 0.741 \\
                     Token length (std) &     text &     0.364 &      0.462 & 0.750 \\
                     Coleman-Liau index &     text &     0.364 &      0.545 & 0.685 \\
                                    RIX &     text &     0.364 &      0.727 & 0.653 \\
                   Sentence Transformer &     text &     0.250 &      0.500 & 0.578 \\
 Proportion Adjacent Dependencies (std) &     text &     0.250 &      0.600 & 0.437 \\
                                    Lix &     text &     0.200 &      0.667 & 0.674 \\
\bottomrule
\end{tabular}
\end{adjustbox}
\end{table*}

\begin{table*}
\caption{Binary performances for ASD models}
\label{tab:asd_perf}
\begin{adjustbox}{height=10cm, center}
\setlength{\tabcolsep}{2.8pt}
\begin{tabular}{rc||ccc}
\toprule
\multicolumn{1}{c}{} & \multicolumn{1}{c}{} & \multicolumn{3}{c}{F1 - ASD} \\
                                        &&     Test & Validation & Train \\
\midrule
                                Alvenir &    audio &    0.750 &      0.706 & 0.716 \\
                            Gjallarhorn &    audio &    0.667 &      0.667 & 0.701 \\
                                  XLS-R &    audio &    0.667 &      0.667 & 0.701 \\
                   Transformer Ensemble & ensemble &    0.667 &      0.444 & 0.810 \\
                        TF-IDF, 10000-d &     text &    0.667 &      0.462 & 0.667 \\
              Dependency distance (std) &     text &    0.667 &      0.400 & 0.667 \\
                                  Dummy & baseline &    0.667 &      0.667 & 0.581 \\
                               FastText &     text &    0.667 &      0.444 & 0.941 \\
            Automated readability index &     text &    0.615 &      0.545 & 0.824 \\
                   Number of characters &     text &    0.600 &      0.500 & 0.706 \\
                   Number unique tokens &     text &    0.600 &      0.500 & 0.750 \\
                Proportion unique okens &     text &    0.571 &      0.500 & 0.667 \\
 Proportion Adjacent Dependencies (std) &     text &    0.545 &      0.615 & 0.444 \\
                   Sentence Transformer &     text &    0.545 &      0.444 & 1.000 \\
                                    RIX &     text &    0.545 &      0.462 & 0.667 \\
                       Number of tokens &     text &    0.545 &      0.500 & 0.706 \\
                    Flesch reading ease &     text &    0.545 &      0.545 & 0.889 \\
                     Coleman-Liau index &     text &    0.533 &      0.571 & 0.667 \\
                      Mean token length &     text &    0.500 &      0.727 & 0.526 \\
                  Sentence length (std) &     text &    0.462 &      0.462 & 0.636 \\
                             ScandiBERT &     text &    0.462 &      0.444 & 0.778 \\
                       All descriptives &     text &    0.462 &      0.500 & 0.667 \\
                            XLM-RoBERTa &     text &    0.444 &      0.500 & 1.000 \\
                                  MFCCs &    audio &    0.429 &      0.769 & 0.853 \\
                                    Lix &     text &    0.400 &      0.364 & 0.625 \\
                      Gunning Fog index &     text &    0.400 &      0.444 & 0.625 \\
             Dependency distance (mean) &     text &    0.400 &      0.545 & 0.667 \\
              Syllables per token (std) &     text &    0.364 &      0.727 & 0.778 \\
                 Sentence length (mean) &     text &    0.364 &      0.667 & 0.800 \\
                   Flesch-Kincaid grade &     text &    0.364 &      0.364 & 0.842 \\
                         TF-IDF, 1000-d &     text &    0.364 &      0.400 & 0.700 \\
                          TF-IDF, 100-d &     text &    0.333 &      0.364 & 0.737 \\
Proportion Adjacent Dependencies (mean) &     text &    0.333 &      0.533 & 0.667 \\
                                Ælectra &     text &    0.286 &      0.500 & 1.000 \\
             Syllables per token (mean) &     text &    0.200 &      0.545 & 0.462 \\
                                eGeMAPS &    audio &    0.200 &      0.727 & 0.688 \\
                     Token length (std) &     text &    0.182 &      0.615 & 0.471 \\
                       Number sentences &     text &    0.000 &      0.000 & 0.000 \\
                                ComParE &    audio &    0.000 &      0.000 & 0.000 \\
                      Sentence polarity &     text &    0.000 &      0.000 & 0.000 \\
                      Baseline Ensemble & ensemble &    0.000 &      0.000 & 0.000 \\
\bottomrule
\end{tabular}
\end{adjustbox}
\end{table*}

\begin{figure*}[!ht]
    \centering
    \includegraphics[width=\textwidth]{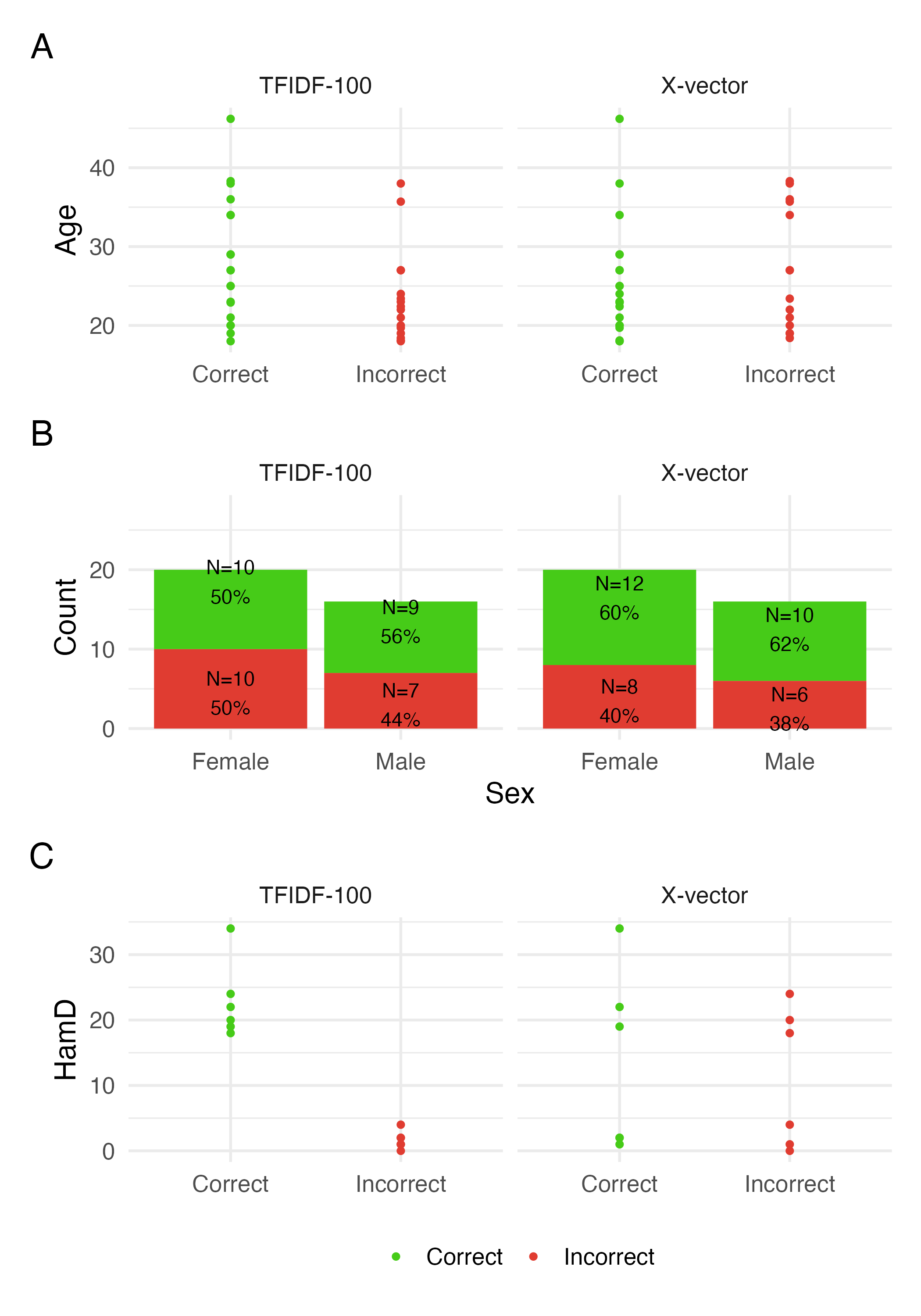}
    \caption{Performance of the X-vector and TFIDF-100 model on the test set stratified by sex, age, and Hamilton score.}
    \label{fig:error_analysis}
\end{figure*}

\end{document}